\documentclass{article}
\usepackage[final,nonatbib]{neurips_2024}
\usepackage[utf8]{inputenc} 
\usepackage[T1]{fontenc}    
\usepackage{hyperref}       
\usepackage{url}            
\usepackage{booktabs}       
\usepackage{nicefrac}       
\usepackage{microtype}      
\usepackage{xcolor}         
\usepackage{listings}%
\usepackage{colortbl}
\usepackage{makecell}
\usepackage{subfigure}
\usepackage{graphicx}%
\usepackage{multirow}%
\usepackage{amsmath,amssymb,amsfonts}%
\usepackage{amsthm}%
\usepackage{mathrsfs}%
\usepackage{textcomp}%

\newcommand{\et}{\emph{et al.}\ }

\title{YOLO-TLA: An Efficient and Lightweight Small Object Detection Model based on YOLOv5}

\author{
Chun-Lin Ji$^{1}$, Tao Yu$^{1}$, Peng Gao$^{1}$, Fei Wang$^{2}$, Ru-Yue Yuan\\
$^1$Qufu Normal University \quad
$^2$Harbin Institute of Technology Shenzhen
}

\begin{document}

\maketitle

\begin{abstract}
Object detection, a crucial aspect of computer vision, has seen significant advancements in accuracy and robustness. Despite these advancements, practical applications still face notable challenges, primarily the inaccurate detection or missed detection of small objects. Moreover, the extensive parameter count and computational demands of the detection models impede their deployment on equipment with limited resources. In this paper, we propose YOLO-TLA, an advanced object detection model building on YOLOv5. We first introduce an additional detection layer for small objects in the neck network pyramid architecture, thereby producing a feature map of a larger scale to discern finer features of small objects. Further, we integrate the C3CrossCovn module into the backbone network. This module uses sliding window feature extraction, which effectively minimizes both computational demand and the number of parameters, rendering the model more compact. Additionally, we have incorporated a global attention mechanism into the backbone network. This mechanism combines the channel information with global information to create a weighted feature map. This feature map is tailored to highlight the attributes of the object of interest, while effectively ignoring irrelevant details. In comparison to the baseline YOLOv5s model, our newly developed YOLO-TLA model has shown considerable improvements on the MS COCO validation dataset, with increases of 4.6\% in mAP@0.5 and 4\% in mAP@0.5:0.95, all while keeping the model size compact at 9.49M parameters. Further extending these improvements to the YOLOv5m model, the enhanced version exhibited a 1.7\% and 1.9\% increase in mAP@0.5 and mAP@0.5:0.95, respectively, with a total of 27.53M parameters. These results validate the YOLO-TLA model's efficient and effective performance in small object detection, achieving high accuracy with fewer parameters and computational demands.
\end{abstract}

\section{Introduction}\label{sec:1}

The rapid development of deep learning in recent years has led to significant breakthroughs in various aspects of computer vision, notably in object detection. This crucial aspect of computer vision aims to identify and classify objects (e.g., pedestrians, animals, vehicles) within images, serving as a foundational element for tasks like object tracking and object segmentation~\cite{r1,r2}. Its industrial applications are extensive, ranging from flaw detection to autonomous driving~\cite{r4,r5,r6}. Furthermore, the evolution of electronic control systems and aircraft design has highlighted the importance of unmanned aerial vehicles (UAV)-based object detection, which has become increasingly prevalent in fields like agriculture, disaster management, and aerial photography. UAVs are either radio-controlled or operate on pre-programmed routes. Equipped with high-resolution cameras, UAVs are capable of capturing comprehensive digital images, facilitating real-time object detection in-flight using lightweight models~\cite{r7,r8,r9}.

Currently, there are two principal methodologies for implementing object detection models: the two-stage and the single-stage methods. The two-stage detection process starts with extracting image features using a convolutional neural network (CNN), followed by the generation of multiple candidate bounding boxes (or regions) on the feature map. Each box is then processed by additional convolutional layers for object classification and bounding box refinements. This includes identifying the object class and verifying its presence in the bounding box, and a regression phase to enhance the detection accuracy. The process concludes with non-maximum suppression (NMS), which filters out surplus bounding boxes to identify the most confident and accurate final detections. In contrast, the single-stage detection method directly predicts object class and their locations in the image, foregoing candidate bounding boxes generation. This method is generally faster and less computationally intensive than the two-stage method, making it ideal for scenarios requiring high real-time performance.

Since its introduction, the YOLO series of object detection methods has garnered significant attention in the computer vision community, with several iterations enhancing its performance in both academic and industrial settings~\cite{r10,r11,r12}. The detection model, YOLOv5, proposed by Ultralytics Inc., boasts improved accuracy and a more streamlined network architecture~\cite{rev1,rev2,rev3,rev4}. Despite the advancements in YOLOv5, including its balance of speed and accuracy, it confronts specific challenges in industrial applications~\cite{r3}. The model often struggles with accurately predicting the positions and classes of small or dense objects, negatively impacting detection performance. Furthermore, deploying these models in embedded systems, which typically have limited resources, necessitates a reduction in the number of parameters and computational demands. Consequently, optimizing the model to be more resource-efficient while preserving accuracy is a critical issue to be addressed. Ultralytics Inc. recognized these issues following the release of the YOLOv5 model and introduced an improved version, YOLOv8~\cite{rev5,rev6,rev7,rev8}. Both generations share similar inference processes and have the same backbone and neck network architectures. However, YOLOv8 introduces two major innovations compared to YOLOv5. Firstly, YOLOv8 employs a more efficient and lighter C2f module in the network architecture, corresponding to the C3 module in YOLOv5. The C2f retains the residual module and adds connections between different scale feature layers. Additionally, the split method is used to replace the convolution operation in the original residual branch. These improvements enable the C2f module to achieve richer gradient flows, enhance feature extraction capabilities, and reduce parameters. Secondly, YOLOv8 replaces the coupled architecture with the currently popular decoupled architecture in the head network and transitions the detection head from anchor-based to anchor-free. These modifications refine the detection process and improve the efficiency and accuracy.

In this paper, we present the YOLO-TLA object detection model based on the YOLOv5 model, which achieves higher accuracy and robustness while maintaining moderate model complexity in terms of parameters and computations. A distinctive improvement of YOLO-TLA is the tiny detection layer in the neck network, enhancing the multi-scale feature fusion process. This allows the model to better focus on small objects in the image. Additionally, YOLO-TLA employs a backbone network enhancement strategy that incorporates attention mechanisms at four locations in the backbone network to improve feature extraction capabilities. By comparing several mainstream attention mechanisms integrated with the YOLOv5s model and applying the best-performing global attention mechanism (GAM)~\cite{r13} to the YOLO-TLA backbone network. Although these improvements positively impact detection capabilities, they inevitably increase the parameters and computational costs. Inspired by the C2f module design in the YOLOv8 model, we enhanced the feature extraction capabilities of the internal modules and simplified their complexity to improve model performance effectively. Specifically, we replace some C3 modules in the YOLOv5 network with lightweight feature extraction modules, C3CrossConv and C3Ghost, proposing four lightweight strategies and exploring their impact on model detection performance and model size. In summary, this study makes the following contributions:
\begin{itemize}
    \item Implementing a tiny detection layer in the neck network of the YOLOv5s model to enhance its tiny object detection performance.
    \item Embedding the C3CrossCovn module into the backbone network of the YOLOv5s model, aiming to streamline the model by reducing its parameters and computational demands.
    \item Comparing the impact of seven current mainstream attention mechanisms on the detection performance and complexity of the YOLOv5s model.
    \item Adding the global attention mechanism (GAM)~\cite{r13} at multiple points in the backbone network, enhancing focus on critical feature information and improving feature extraction efficiency.
    \item Applying the aforementioned optimizations and improvements to the YOLOv5s model, proposing the final optimized model YOLO-TLAs.
    \item Extending the design method of YOLO-TLA to the larger YOLOv5m model, proposing the YOLO-TLAm model.
\end{itemize}

The remainder of this paper is organized as follows. Existing object detection methods are introduced in Section~\ref{sec:2}. The following section, Section~\ref{sec:3}, proposes a small object detection model based on YOLOv5. In Section~\ref{sec:4}, experimental results and discussions demonstrate that the proposed method is efficient for small object detection. Finally, a brief conclusion is provided in Section~\ref{sec:5}.
\section{Related works}\label{sec:2}

\subsection{Object detection}

The field of deep learning-based object detection has primarily focused on two approaches: CNN-based and Transformer-based methods.

R-CNN~\cite{r14}, developed by Girshick \et, was a significant step, using a pre-trained CNN for feature extraction in various candidate regions and SVMs for classification. Despite its innovation, R-CNN suffered from redundant and complex processing, consuming extensive computational resources and storage.
This led to the development of Fast R-CNN~\cite{r15}, which introduced the region proposal network (RPN) to replace the selective search approach. RPN generates candidate regions directly from the feature map, minimizing duplicated computations, and thereby enhancing model efficiency and speed. Moreover, Fast R-CNN employs a single feature extraction network for both classifying candidate boxes and performing bounding box regression, improving the use of features and detection accuracy.
Advancing further, Ren \et proposed Faster R-CNN~\cite{r16}, a landmark model in object detection. Faster R-CNN innovatively integrates region proposal generation with the feature extraction network, enabling comprehensive end-to-end training covering region proposal, feature extraction, classification, and bounding box regression.
YOLO~\cite{r17}, proposed by \textcolor{blue}{Redmon} \et, approaches object detection by initially partitioning the input image into several grids. Each grid predicts potential bounding boxes for objects, spanning the entire image. These predictions are then refined through NMS to derive the final detection results.
Liu \et developed SSD~\cite{r18}, enhancing the VGG16 network architecture for multi-scale feature map detection. SSD leverages this enhanced backbone network to extract feature maps at various levels, each with distinct scales and features. Detection and classification are performed on these maps, and their predictions are aggregated to form the final detection outcome.
YOLOv3~\cite{r19}, proposed by Redmon \et, marks a significant advancement by replacing its feature extraction network with DarkNet53, a network architecture known for its depth and superior feature extraction capabilities. YOLOv3 also leverages the feature pyramid network (FPNet)~\cite{r37}, using multi-scale downsampled feature maps for enhanced detection of small objects, thereby improving both inference speed and accuracy.
On the other hand, Duan \et introduced CenterNet~\cite{r20}, which eschews NMS and RPN, reducing accuracy degradation. CenterNet focuses on detecting object centers and calculating final bounding boxes with corner points and offset guidance.
Tan \et then proposed EfficientDet~\cite{r21}, which incorporates EfficientNet as its backbone network and introduces a bi-directional weighted feature pyramid network (BiFPN). The BiFPN allows for better feature representation through bidirectional cross-scale connectivity and efficient fusion, significantly enhancing detection speed while maintaining accuracy.
Bochkovskiy \et introduced YOLOv4~\cite{r10}, a detection model that upgrades its backbone network to CSPDarknet53, an enhanced version of DarkNet53 featuring a CSP module. This enhancement results in a more modular network simplifies the backbone network architecture, and augments feature extraction capabilities. YOLOv4 further evolves from YOLOv3 by adding a path aggregation network (PAN) architecture to its neck, drawing from the principles of PANet~\cite{r38} to enhance multi-scale feature fusion.
Ge \et proposed YOLOx~\cite{r11}, building upon YOLOv5 while maintaining much of its design philosophy. YOLOv5 excels in single-stage object detection with its high accuracy and swift detection capabilities. YOLOx modifies YOLOv5 by implementing a decoupled detection head, separating the predictions of object class, bounding box, and intersection ratio into distinct branches. This decoupling has been found to not only enhance detection accuracy but also speed up model convergence.
On a different note, Dosovitskiy \et pioneered the application of Transformer in computer vision with the Vision Transformer (ViT)~\cite{r24}, revolutionizing the research field of object detection. ViT adopts an encoder-decoder and self-attention mechanism to capture global image features, thus enabling complete end-to-end detection. However, this innovation comes at the cost of increased model parameters, raising training and deployment challenges.
Concurrently, Carion \et introduced DETR~\cite{r25}, the first to apply the Transformer architecture to object detection. DETR consists of a CNN-based backbone, an encoder, a decoder, and a feed forward network, forming a comprehensive system for detection. Despite its high parameter count and limited accuracy enhancements, DETR has been a cornerstone for subsequent advancements in Transformer-based object detection.
Gold-YOLO~\cite{r26}, proposed by Wang \et, represents a departure from the conventional YOLO model improvements. It introduces the gather-and-distribute mechanism to the neck network, significantly boosting the feature fusion capabilities and enabling more efficient information interaction.
In a different approach, Chen \et developed DiffusionDet~\cite{r27}, a pioneering effort in applying diffusion approaches to object detection. It redefines the detection task as a stepwise process of bounding box generation, utilizing decoupled training and iterative dynamic box evaluation.
Zheng \et introduced Focus-DETR~\cite{r28}, a lightweight modification of DETR. Focus-DETR addresses the high complexity and parameter count that DETR faces by selectively processing object feature vectors, prioritizing only key objects. This selective approach can be tailored by adjusting the number of input vectors.

Object detection methods based on CNNs typically utilize deep and wide backbone networks for feature extraction. They leverage multi-scale feature fusion~\cite{r29} to capture extensive semantic information without neglecting geometric texture details, thereby enhancing the expressiveness of the detection feature map. However, these methods come with a trade-off, as the numerous convolutional operations and stacked feature extraction networks significantly increase the model complexity and parameter count.
On the other hand, Transformer-based object detection methods utilize self-attention mechanisms to model the interrelationships of different feature maps and globally contextualize them. These methods are adept at assimilating information from multi-scale receptive fields. However, the complexity inherent in the Transformer architecture, particularly the interactions among several feature vectors, results in an increased number of parameters and higher computational load. Moreover, challenges such as object occlusion and background distraction persist in detection tasks.
Therefore, current research primarily focuses on enhancing detection accuracy and robustness, achieving real-time processing, and reducing model complexity. This paper aims to bolster the detection of small objects by augmenting the feature extraction capabilities of the backbone network and the multi-scale feature fusion in the neck network. Simultaneously, we intend to decrease the overall parameters and computational demands through the application of lightweight strategies.

\subsection{Attention mechanisms}

Inspired by the way humans perceive visual information, attention mechanisms in computer vision aim to emulate selective focus on objects, minimizing attention to backgrounds and distractions. Attention mechanisms are broadly classified into two types: channel attention and spatial attention.

One notable effort is the squeeze-and-excitation network (SENet)~\cite{r30}. It consists of two phases: the squeeze phase, which assesses feature distribution across each channel-wise feature map, and the excitation phase, which leverages this distribution to discern the dependencies between channels and assign appropriate weights to each. By focusing on channel-specific feature weighting and dependencies, SENet effectively concentrates on region of interests (ROIs), offering an efficient alternative to depth network architectures with its minimal computational and parameter demands.
Woo \et introduced the convolutional block attention module (CBAM)~\cite{r31}, merging spatial attention module (SAM) with channel attention module (CAM) to augment object detection capabilities. While CAM focuses on enhancing channel features related to the object, SAM is designed to capture spatial information, thereby boosting the ability to understand spatial relationships.
Li \et introduced selective kernel network (SKNet)~\cite{r32}, which employs selectivity coefficients to dynamically adjust convolution kernel sizes, capturing multi-scale features effectively in complex scenes.
Building on SENet, Wang \et proposed efficient channel attention (ECA)~\cite{r33}, which optimizes channel feature weighting using global average pooling and one-dimensional convolution. ECA also dynamically calculates the convolution kernel size based on the feature map channel count, striking a balance between computational efficiency and performance.
Coordinate attention (CA)~\cite{r34}, proposed by Hou \et, concentrates on spatial coordinate information in feature maps. By learning coordinate weights, CA modifies feature representations across different locations, enhancing the spatial understanding and generalization abilities of the network.
Zhang \et developed the shuffle-attention (SA)~\cite{r35}, which segments input feature maps into groups along the channel dimension. Each group passes through a channel branch for generating feature weights via global average pooling, and a spatial branch for spatial statistical computation. These branch outputs are then combined to enhance subgroup integration. Despite its effectiveness, SA incurs a significant increase in parameters and computational load due to its grouping and aggregation process.
Additionally, Zhang \et proposed RFACovn~\cite{r36}, which fuses standard convolution with spatial attention. This method calculates the weights for each region within a receptive field through spatial attention, assigning unique convolution kernels to each region.

Existing attention mechanisms in computer vision, encompassing channel and spatial attention or their combination, typically enhance object representation via feature weighting. These mechanisms are advantageous as they amplify important features without significantly adding to model complexity, thanks to their straightforward modular architecture. Our study builds on this philosophy, integrating attention mechanisms at various positions within the backbone network. Unlike conventional methods that focus solely on the channel or spatial dimensions, our mechanism also captures the global features of the image, offering a more comprehensive understanding of the input feature maps.

\section{Methodology}\label{sec:3}

\subsection{Motivation and baseline}
\begin{figure}[t!]
    \begin{center}
        \includegraphics[width=\linewidth]{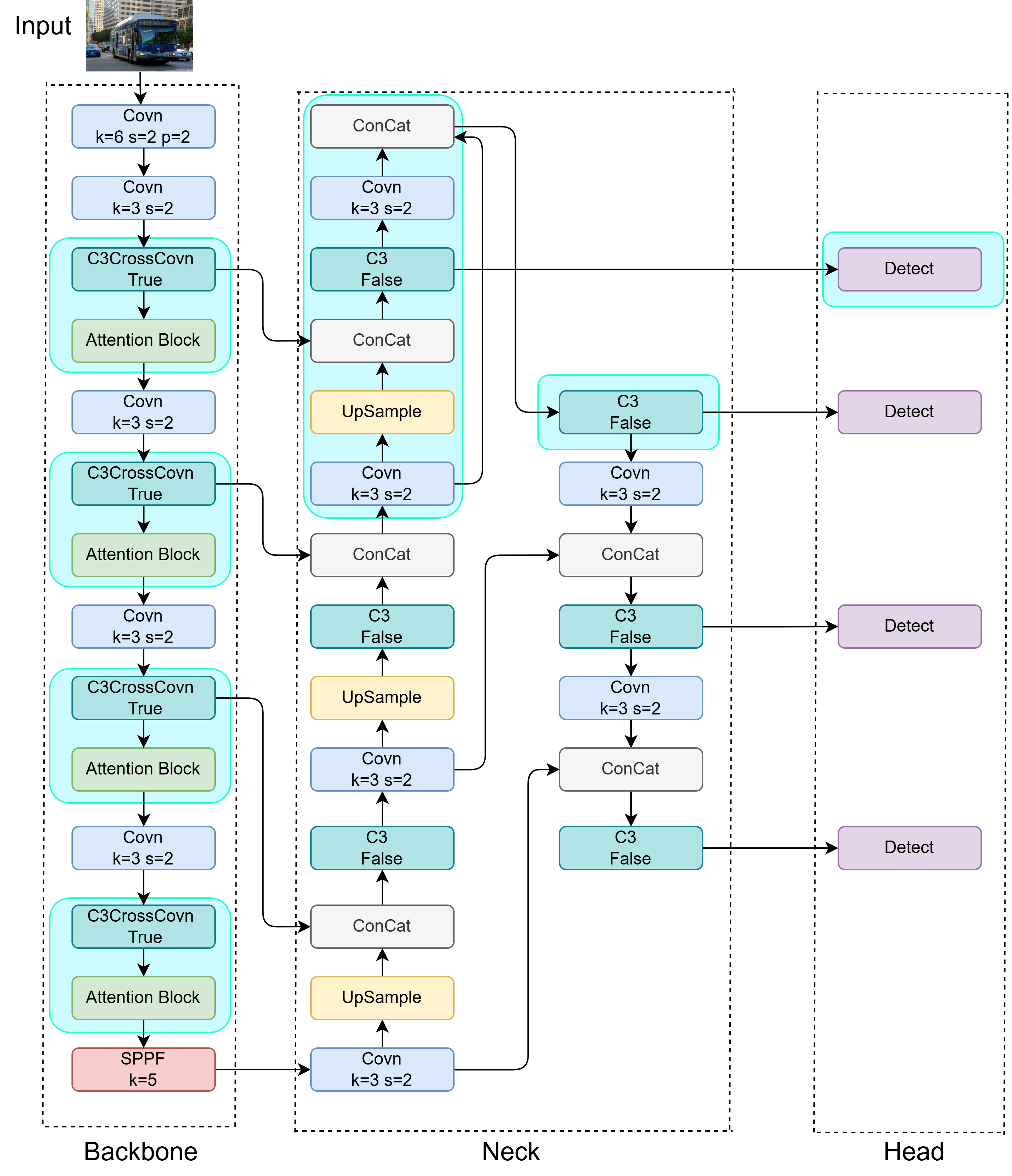}
    \end{center}
    \caption{Overview network architecture of the proposed YOLO-TLA. In the figure, $k$, $s$, and $p$ indicate the convolutional kernel size, stride, and padding size, respectively. The light green regions show our main improvements to the baseline method YOLOv5.}
    \label{fig:1}
\end{figure}

In this study, we propose YOLO-TLA, an improved object detection model based on YOLOv5, with a focus on small object detection and reduced model complexity, as outlined in Fig.\ref{fig:1}. YOLOv5 presents five versions, named YOLOv5n, YOLOv5s, YOLOv5m, YOLOv5l, and YOLOv5x, in order of increasing size. Each version is designed with specific configurations to cater to the needs of different sizes. The model is structured into three primary parts: the backbone network, the neck network, and the head network. The backbone network is built on CSPDarknet53, consisting of standard convolutional layers with additional feature enhancement modules, tasked with extracting geometric texture features like the shape and color of objects. To enrich this basic information, the neck network, drawing inspiration from FPNet~\cite{r37} and PANet~\cite{r38}, further combines feature maps from the backbone network with deeper semantic information. This combination results in feature maps rich in both semantic and geometric information. These enhanced feature maps are then fed into the head network, which performs the final detection and classification.

\subsection{Tiny object detection layer}

In the MS COCO dataset~\cite{r39}, objects are categorized based on size: small objects are less than 32$\times$32 pixels, medium objects range from 32$\times$32 to 96$\times$ 96 pixels, and large objects exceed 96$\times$96 pixels. To enhance small object detection in our model, we have adjusted the feature map and anchor sizes. Utilizing the k-means clustering algorithm, we recalibrated the size range of preset anchor boxes and introduced a tiny object detection layer, leading to the creation of the YOLOv5s-Tiny model. Specifically, we upsample in the neck network of YOLOv5s to produce a 160$\times$160 feature map with 128 channels, which is then combined with the third layer output of the backbone network, matching in channels and size. This combined feature map, along with other detection layer outputs, is processed in the head network for classification and detection. Our neck network generates several feature maps at different scales, each corresponding to various anchor sizes. This enables the detection of smaller objects on larger feature maps and larger objects on smaller feature maps, enhancing object representation in the image. The correlation between feature map sizes and anchor sizes is detailed in Table \ref{tab:1}.
\begin{table}[t!]
\centering
\caption{Anchor box sizes in different feature maps.}
\label{tab:1}
\begin{tabular}{cc}
\toprule
Size of feature maps & Sizes of anchor boxes     \\
\midrule
160$\times$160                  & 9$\times$12, 20$\times$19, 17$\times$42           \\
80$\times$80                    & 43$\times$26, 36$\times$56, 76$\times$52          \\
40$\times$40                    & 49$\times$121, 108$\times$102, 111$\times$121     \\
20$\times$20                    & 231$\times$138, 230$\times$325, 479$\times$372    \\
\bottomrule
\end{tabular}%
\end{table}

\subsection{Lightweight convolution module}

To streamline the YOLOv5 model by reducing its parameter count and computational demands, this study explores the integration of two modules: C3Chost and C3CrossCovn. We aim to assess the impact of these modules when inserted into different positions within the backbone network. This comparison focuses on how each module placement affects the performance and complexity of the model, with the goal of simplifying the network architecture without compromising its performance.

\subsubsection{C3 module revisited}

\begin{figure}[t!]
    \begin{center}
        \includegraphics[width=\linewidth]{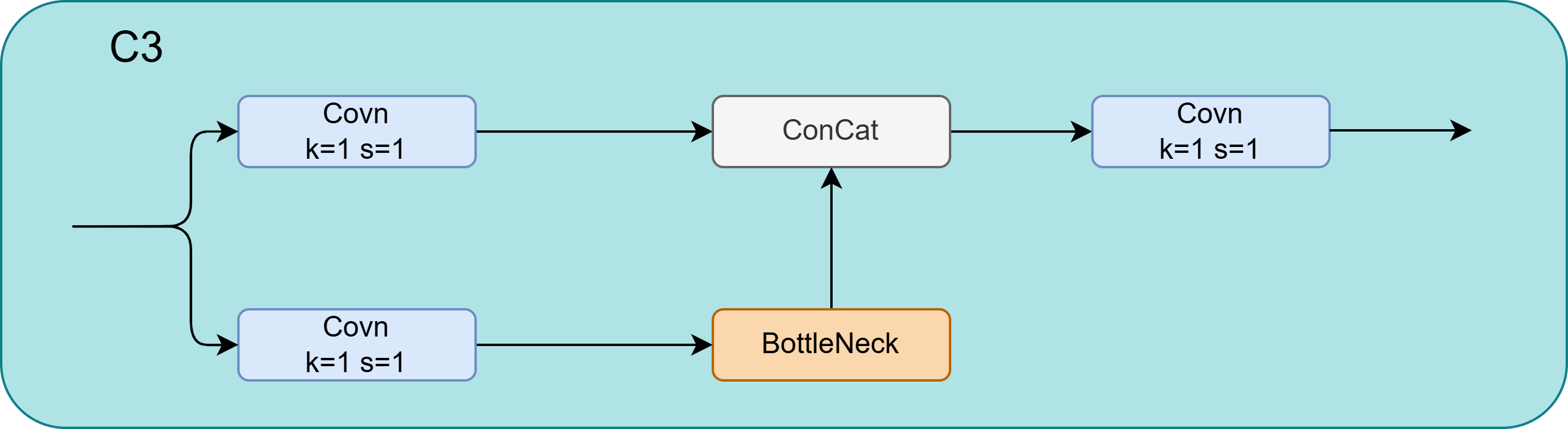}
    \end{center}
    \caption{Overview of the C3 module. In the figure, $k$ and $s$ indicate the convolutional kernel size and stride, respectively.}
    \label{fig:2}
\end{figure}

The C3 module, a key component of YOLOv5, shown in Fig.\ref{fig:2}, comprises three standard convolutional layers, each with a 1$\times$1 kernel size and stride of 1, and includes several stacked BottleNeck modules. The architecture of this module varies in width and depth depending on the model size, controlled by predefined parameters. The C3 module incorporates a residual structure akin to BottleNeckCSP. It processes the input feature map in two ways: either through a dual-branch approach, where a single branch uses two standard convolutional layers and the other outputs the original feature map, subsequently concatenating the outputs, or by forgoing the residual path and directly outputting the feature map after standard convolutions. The BottleNeck module in C3 is noted for its robust feature extraction capabilities and its role in addressing gradient vanishing and explosion challenges.

\subsubsection{C3Ghost module}

Standard convolution modules, typically comprising a regular convolution layer along with batch normalization and an activation function, often create numerous similar feature maps, leading to high computational demands and resource consumption. To address this, ghost convolution (GhostCovn) adopts a two-step approach. Initially, it employs a standard convolution with a smaller kernel size to generate a feature map with fewer channels. Subsequently, depthwise convolution (DepConv) is used to produce the additional feature maps that were not created in the first step. These feature maps from both phases are then combined, yielding a final feature map akin to that produced by a standard convolution layer but with considerably less computation and fewer parameters.
\begin{figure}[t!]
    \begin{center}
        \includegraphics[width=\linewidth]{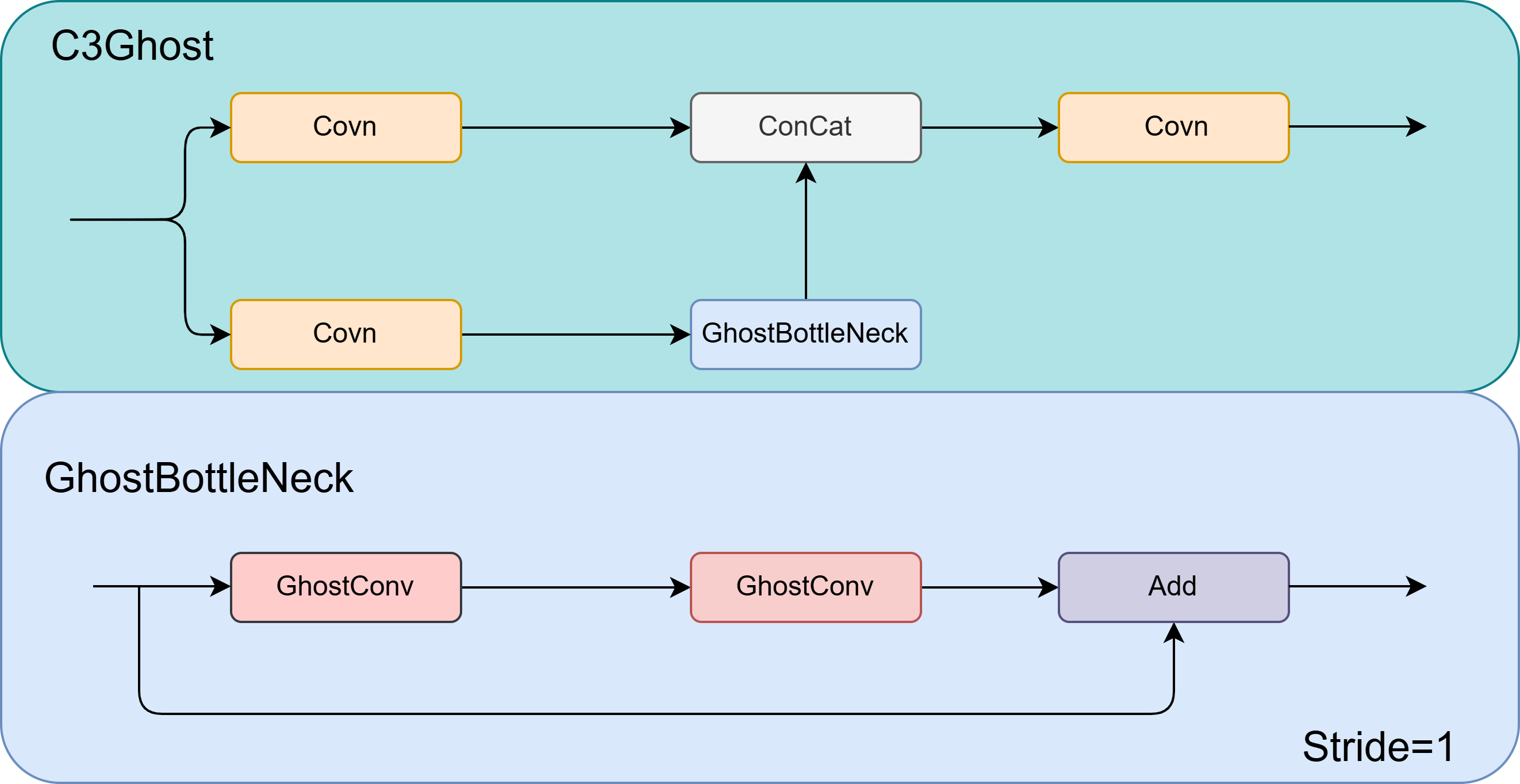}
    \end{center}
    \caption{The pipeline of the C3Ghost module.}
    \label{fig:3}
\end{figure}

In our proposed C3Chost module, the GhostCovn is utilized in conjunction with the GhostBottleNeck module, which includes both GhostCovn and DepConv modules and features a residual structure. The GhostBottleNeck comes in two branches: the first involves processing input feature maps through a GhostCovn with 1$\times$1 convolution kernels and a stride of 1, then adding the resultant feature map to the original feature map. The second branch introduces an intermediate DepConv module with a 3$\times$3 kernel and stride of 2 between two GhostCovn modules. The residual paths follow a similar DepConv module, then a standard 1$\times$1 convolution with stride 1. The overall structure of the C3Chost module is akin to the C3 module, but it substitutes the BottleNeckCSP module with the GhostBottleNeck. The overview architecture is illustrated in Fig.\ref{fig:3}.

\subsubsection{C3CrossCovn module}

\begin{figure}[t!]
    \begin{center}
        \includegraphics[width=0.7\linewidth]{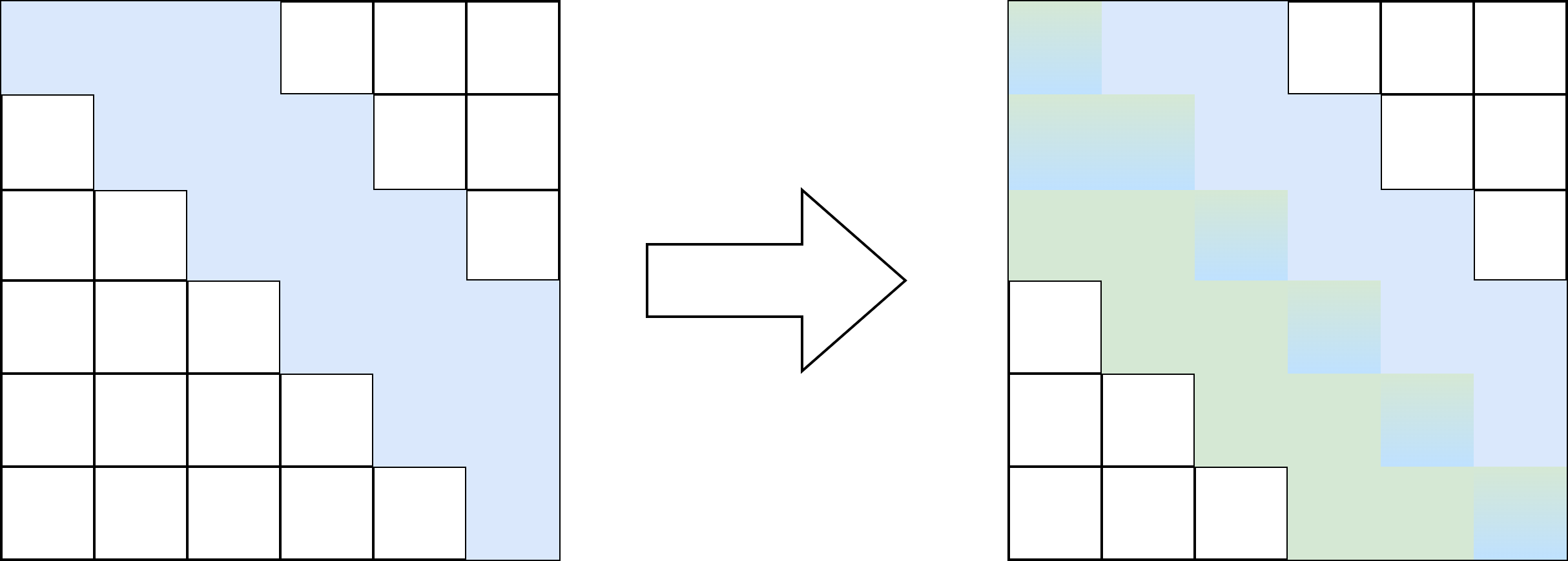}
    \end{center}
    \caption{Illustration of the CrossCovn module with stride of 1 and kernel size of 3}
    \label{fig:4}
\end{figure}

While GhostCovn significantly simplifies the C3 module, it inadvertently leads to a loss of representative information along the channel-wise direction, affecting model accuracy. To mitigate this, cross convolution (CrossCovn) is employed. CrossCovn comprises two standard convolutional layers arranged in a cross pattern over the feature map. It diverges from traditional $k\times k$ sliding window convolutions, utilizing a 1$\times k$ kernel for the first layer with a horizontal stride of 1 and a vertical stride of $s$, and a $k\times$1 kernel for the second layer with equal strides of $s$ in both dimensions. An illustration of CrossCovn is provided in Fig.\ref{fig:4}.

In order to assess the distinction in parameter count and computational load between standard convolution and CrossCovn, we set up a comparative framework. For this analysis, let's consider a square input image with dimensions H$\times$H$\times$C. The convolution employs a $k\times k$ kernel size, and the number of kernels equals the number of channels, $C$, with padding $p$ around the image edges. Eq.\ref{eq:1} determines the floating-point computational demands and the parameter count needed for processing through a standard convolutional layer,
\begin{equation}\label{eq:1}
  \begin{aligned}
    \text{FLOPs}_1 & = k^2C(\frac{W-k+2p}{s}+1)^2 \\
    \text{parameters}_1 & = k^2C
  \end{aligned}
\end{equation}

To determine the computational demands required by CrossCovn for a single operation on an image, we make the following assumptions: CrossCovn contains $C$ dual-convolutional kernels, the first kernels are sized 1$\times k$, and the second kernels are sized $k\times$1, with a stride of $s$. Eq.\ref{eq:2} details the specific calculations for the computational load and the number of parameters associated with these settings in CrossCovn,
\begin{equation}\label{eq:2}
  \begin{aligned}
    \text{FLOPs}_2 & = k^2C(\frac{W-1+2p}{s}+1)(\frac{W-k+2p}{s}+1) \\
    \text{parameters}_2 & = 2kC
  \end{aligned}
\end{equation}

\begin{figure}[t!]
    \begin{center}
        \includegraphics[width=\linewidth]{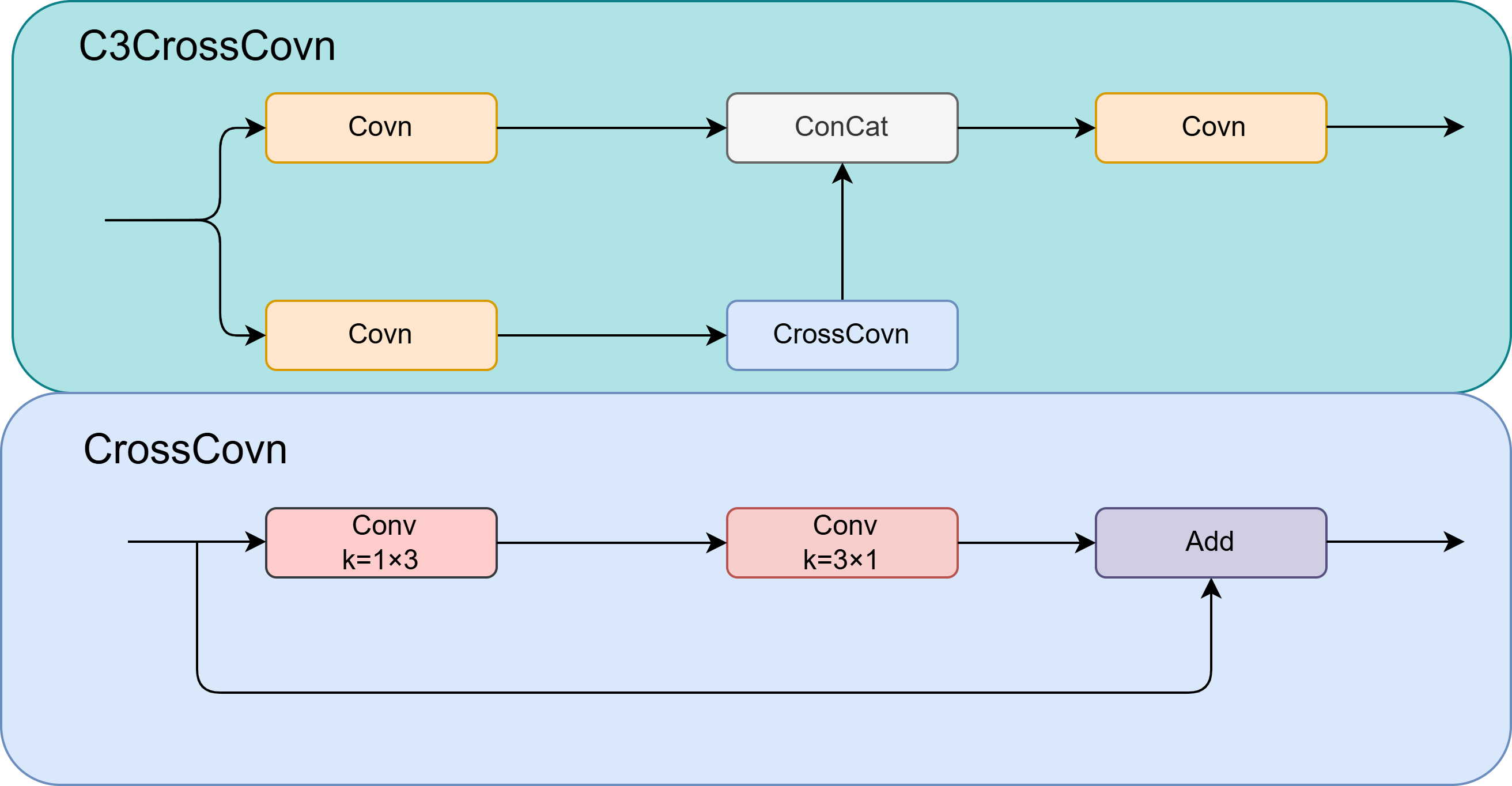}
    \end{center}
    \caption{The pipeline of the C3CrossCovn module.}
    \label{fig:5}
\end{figure}

In the MS COCO dataset, all images are RGB, so we use three image channels ($C=3$). To ensure a sufficiently large receptive field, we set $k$ to 3 and $s$ to 1. Under these conditions, it becomes apparent that standard convolution requires more computational effort and has approximately 1.5 times the number of parameters compared to CrossCovn. Although CrossCovn entails two striped-kernel convolution operations on a single feature map, it achieves finer feature extraction and richer feature information than standard convolution. This enhancement not only increases the detection accuracy but also significantly lowers computational demands and parameter count, making it an optimal solution for model lightweight. The overview of the C3CrossCovn module is depicted in Fig.\ref{fig:5}.

\begin{figure}[t!]
    \begin{center}
        \includegraphics[width=\linewidth]{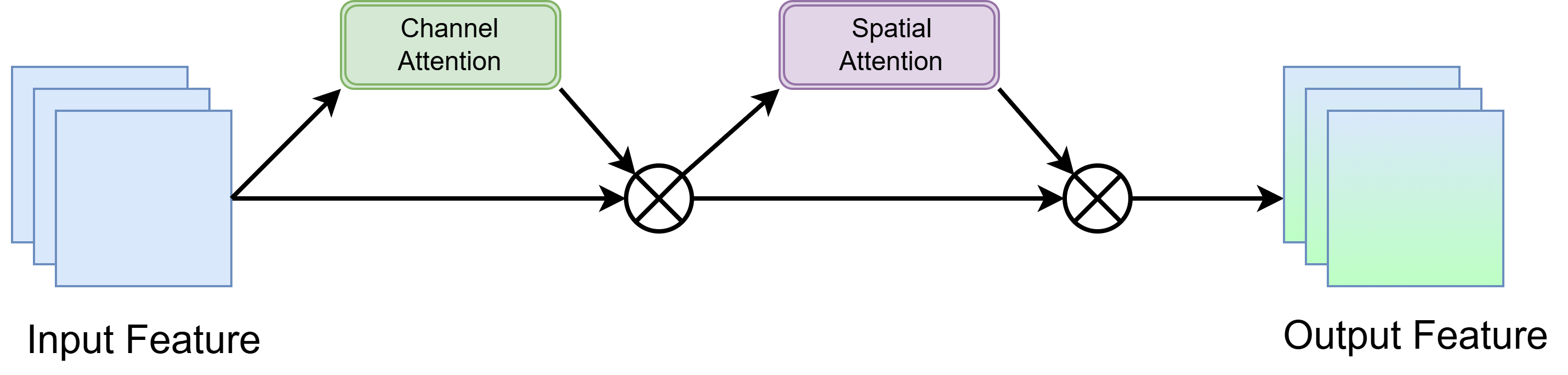}
    \end{center}
    \caption{Overview pipeline of the GAM. The symbol $\bigotimes$ indicates multiplication.}
    \label{fig:6}
\end{figure}

\subsection{Global attention mechanism}

To address the challenges of object detection in computer vision, where objects are often amidst complex backgrounds and distractors, we utilize GAM in various locations within the backbone network. GAM assists the model in focusing more on the objects of interest and reducing interference, thereby improving model accuracy and robustness. Classical attention mechanisms such as SENet and CA have limitations. SE acquires channel correlation weights via the squeeze and excitation steps but neglects spatial information. CA, which employs positional coding, can overemphasize position details, potentially leading to overfitting due to its lack of global correlation analysis. GAM operates in two primary parts: channel attention and spatial attention. This mechanism focuses on both the channel characteristics and the global aspects of the image, using global information to weight the input feature maps for improved accuracy and robustness. Structurally akin to CBAM, GAM sequentially processes the channel attention and then the spatial attention, while also incorporating a residual structure to conserve the feature information of the original image. In channel attention, the feature map undergoes a dimension exchange between channel number $C$ and width W through a permute operation, followed by a multi-layer perceptron (MLP) with two fully-connected layers and a ReLU activation, culminating in a Sigmoid activation. For spatial attention, instead of pooling methods, GAM employs two 7$\times$7 standard convolutions with a ReLU activation. The detailed pipeline of the GAM is depicted in Fig.\ref{fig:6}.

\subsection{Detection pipeline}

In YOLO-TLA, to manage varying input image sizes, all detection images are standardized to 640$\times$640$\times$3 pixels during data augmentation, typically as RGB images. The backbone network extracts features maps at different scales: 320$\times$320$\times$64, 160$\times$160$\times$128, 80$\times$80$\times$256, 40$\times$40$\times$512, and 20$\times$20$\times$1024. The output from this network is a 20$\times$20 feature map with 1024 channels. The neck network, incorporating FPN, deepens feature extraction by fusing feature maps of identical scales and progressively upsampling them to 160$\times$160 with 128 channels. PAN within the neck network further downsamples the feature map to 20$\times$20 with 1024 channels, continuing the feature fusion process. The neck network produces four scales of feature maps (160$\times$160$\times$128, 80$\times$80$\times$256, 40$\times$40$\times$512, and 20$\times$20$\times$1024) for detection. The head network selects anchor boxes based on the size of these feature maps, using sizes derived from clustering the ground-truth annotations in the training dataset, and then classifies objects and regresses anchor boxes to identify their locations and sizes.

\section{Experiments}\label{sec:4}

\subsection{Implement Details}

In the experiments conducted for this study, the MS COCO dataset~\cite{r39} was employed. Specifically, the COCO training set, comprising 118,287 images and 117,266 labels, and the COCO validation set, with 5,000 images and 4,952 labels, were used. Given the varying sizes of these images, they were uniformly resized to 640$\times$640 pixels for the experiments. The COCO training set includes a broad range of objects, totaling 80 object categories, which represent a comprehensive collection of common objects in daily life. This makes the MS COCO dataset widely applicable and valuable for research in computer vision. For training the model, the Adam optimizer is employed, starting with a learning rate of 0.001 and increasing to 0.01. To enhance the speed of parameter updates, we set the momentum to 0.937. Weight decay, crucial for regularization in training, is carefully balanced at 0.0005 to avoid overfitting or underfitting the model. The training regimen includes an initial warm-up phase covering the first 3 epochs, followed by an extensive training period spanning a total of 400 epochs. The model was implemented in Python, utilizing the PyTorch framework, and underwent training on a cloud server outfitted with four 16GB Tesla V100 GPUs. To expedite the experimental process while maintaining result accuracy, we employed different levels of data augmentation tailored to the varying model sizes. In addition, to ensure the accuracy and rigor of the experimental results and to exclude the interference of issues such as experimental hardware and software versions on the experimental results, all models mentioned in the experimental section of this paper, including the YOLOv5 model, were trained, validated, and tested using the same hardware platform and software environment.

\subsection{Evaluation metrics}

Several well-established metrics were chosen for evaluating the model performance in object detection tasks, including precision, recall, F1 score, and mean average precision (mAP), all of which serve to measure the efficacy of detection. Metrics like the count of parameters and floating-point operations (FLOPs) were utilized to assess the complexity of the model. Precision and recall metrics are derived from a confusion matrix, which categorizes predictions into four types based on ground-truth and predicted labels. True positives (TP) occur when both the ground-truth and the prediction are positive. False negatives (FN) happen when the ground-truth is actually positive, but the prediction incorrectly indicates negative. False positives (FP) are cases where the ground-truth is actually negative, but the prediction is incorrectly positive. Finally, True negatives (TN) are instances where both the ground-truth and the prediction are negative. Precision and recall are calculated as follows,
\begin{equation}\label{eq:3}
  \begin{aligned}
    \text{Precision} & = \frac{\text{TP}}{\text{TP}+\text{FP}} \\
    \text{Recall} & = \frac{\text{TP}}{\text{TP}+\text{FN}}
  \end{aligned}
\end{equation}

Given the extensive range of object categories in the MS COCO dataset employed in our experiment, it is impractical to display precision and recall for each category individually. As a result, we presents the metrics as averages across all categories, with average precision and recall represented by Precision$_{all}$ and Recall$_{all}$, respectively. The formulas are expressed as follows,
\begin{equation}\label{eq:4}
  \begin{aligned}
    \text{Precision}_{all} & = \frac{1}{n}\sum_{i=1}^n\text{Precision}_{i} \\
    \text{Recall}_{all} & = \frac{1}{n}\sum_{i=1}^n\text{Recall}_{i}
  \end{aligned}
\end{equation}
where $n$ denotes the total number of categories, and $i$ is the category index.

Usually, precision and recall are inversely related, which suggests that relying on just one of these metrics might not sufficiently evaluate the detection performance. To address this, the F1 score is introduced, serving as a balanced measure. Defined as the harmonic mean of precision and recall, the F1 score encapsulates both aspects in its calculation. The formula to compute the F1 score is outlined below,
\begin{equation}\label{eq:5}
    \text{F1} = \frac{2\times\text{Precision}_{all}\times\text{Recall}_{all}}{\text{Precision}_{all}+\text{Recall}_{all}}
\end{equation}

In object detection, model prediction accuracy is evaluated by comparing the IOU of predictions with the ground-truths. The IOU metric measures how closely the predicted bounding box $\text{B}_{gt}$ aligns with the ground-truth box $\text{B}_{gt}$. Adopting a higher IOU threshold denotes a more rigorous evaluation standard. The specific formula used to calculate IOU is provided below,
\begin{equation}\label{eq:6}
    \text{IOU} = \frac{\text{B}_{pre}\bigcap\text{B}_{gt}}{\text{B}_{pre}\bigcup\text{B}_{gt}}
\end{equation}
Hence, in object detection models, the metrics of precision and recall are derived from the IOU values.

For single-object detection tasks where only one class of object is involved, the overall detection performance of the model can be evaluated using average precision (AP). AP, a stringent measure, assesses the prediction accuracy for all objects of the same class in the dataset, considering a specific IOU threshold. To calculate AP, both precision and recall are considered, forming a precision-recall (P-R) curve, with recall on the x-axis and precision on the y-axis. The value of AP is derived from the area under this curve, and is calculated as follows,
\begin{equation}\label{eq:7}
    \text{AP} = \int_0^1\text{Precision}(\text{Recall})d(\text{Recall})
\end{equation}

For multi-object detection tasks with various object categories, relying solely on AP for each class doesn't provide a complete picture of the model's effectiveness. Mean average precision (mAP) is thus utilized as a more encompassing metric, averaging the AP values for all object classes in the dataset. This results in a more thorough evaluation, accurately reflecting the overall performance of the model. The calculation of mAP is based on the following equation,
\begin{equation}\label{eq:8}
    \text{mAP} = \frac{1}{n}\sum_{i=1}^{n}\text{AP}_i
\end{equation}
where $n$ stands for the total number of categories in the dataset, and $i$ is the index for each category. In this study, mAP@0.5 denotes the mAP calculated at an IOU threshold of 0.5. Furthermore, mAP@0.5:0.95 is the average of mAP values calculated at IOU thresholds that start at 0.5 and increase by increments of 0.05, up to 0.95.

\subsection{Evaluations on tiny object detection layer}

\begin{table*}[t!]
\centering
\caption{Detection results of tiny object detection layers.}
\label{tab:2}
\resizebox{\linewidth}{!}{%
\begin{tabular}{lccccccc}
\toprule
Models & Precision$_{all}$ & Recall$_{all}$ & mAP@0.50 & mAP@0.5:0.95 & F1 & GFLOPs & Parameters (M) \\
\midrule
YOLOv5s~\cite{rev1} & 67.7\% & 50.3\% & 55.7\% & 36.1\% & 0.577 & \textbf{16.6} & \textbf{7.23} \\
YOLOv5s-Tiny & \textbf{68.7\%} & \textbf{52.5\%} & \textbf{57.5\%} & \textbf{37.5\%} & \textbf{0.595} &19.9 &7.38 \\
\bottomrule
\end{tabular}%
}
\end{table*}

\begin{figure*}[t!]
\begin{center}
    \subfigure[]
    {\label{fig:7a}\includegraphics[width=0.33\linewidth]{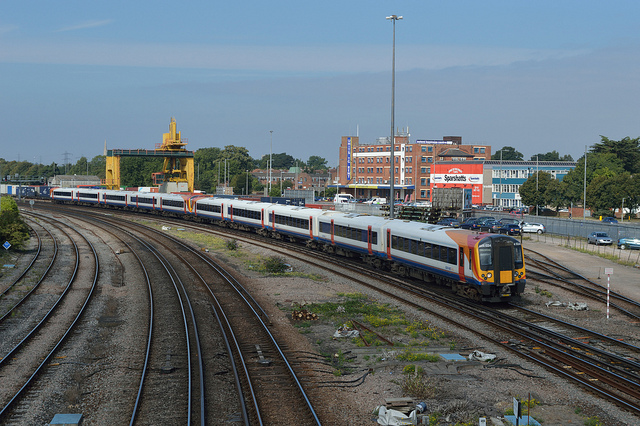}}
    \hfill
    \subfigure[]
    {\label{fig:7b}\includegraphics[width=0.33\linewidth]{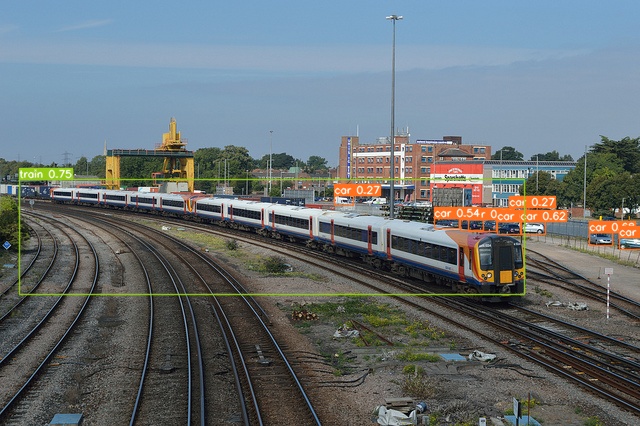}}
    \hfill
    \subfigure[]
    {\label{fig:7c}\includegraphics[width=0.33\linewidth]{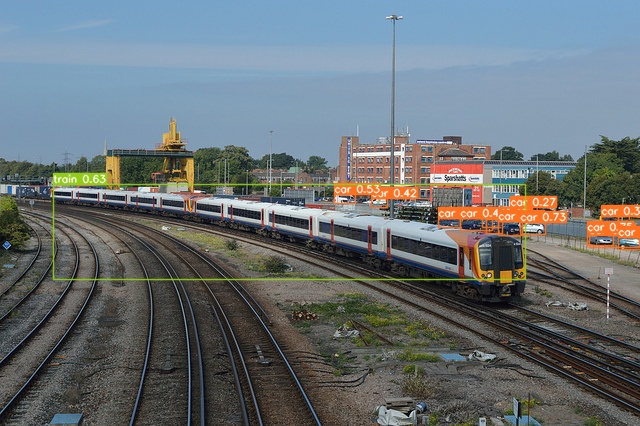}}
\end{center}
\caption{Detection results of YOLOv5s and YOLOv5s-Tiny.}
\label{fig:7}
\end{figure*}

\begin{figure*}[t!]
\begin{center}
    \subfigure[]
    {\label{fig:8a}\includegraphics[width=0.33\linewidth]{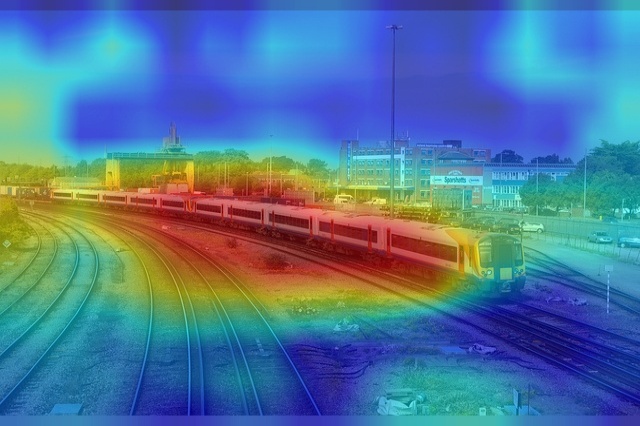}}
    \hfill
    \subfigure[]
    {\label{fig:8b}\includegraphics[width=0.33\linewidth]{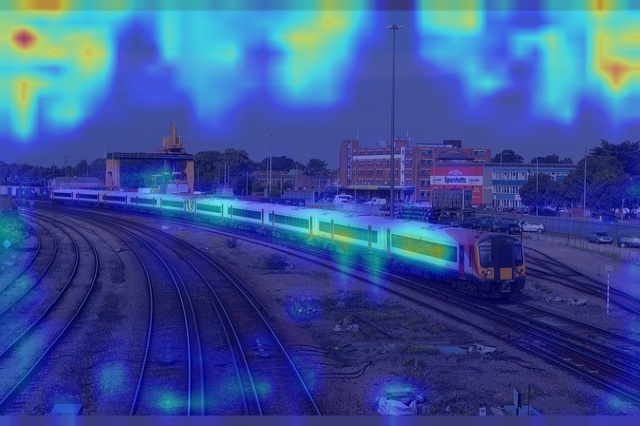}}
    \hfill
    \subfigure[]
    {\label{fig:8c}\includegraphics[width=0.33\linewidth]{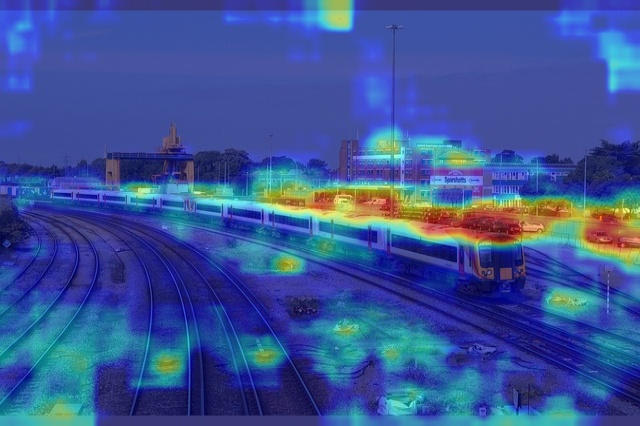}}
\end{center}
\caption{Output heatmaps of YOLOv5s neck network.}
\label{fig:8}
\end{figure*}

\begin{figure*}[t!]
\begin{center}
    \subfigure[]
    {\label{fig:9a}\includegraphics[width=0.33\linewidth]{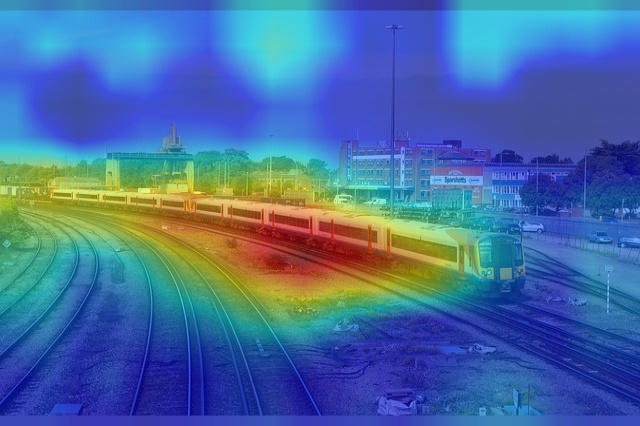}}
    \hspace{0.05em}
    \subfigure[]
    {\label{fig:9b}\includegraphics[width=0.33\linewidth]{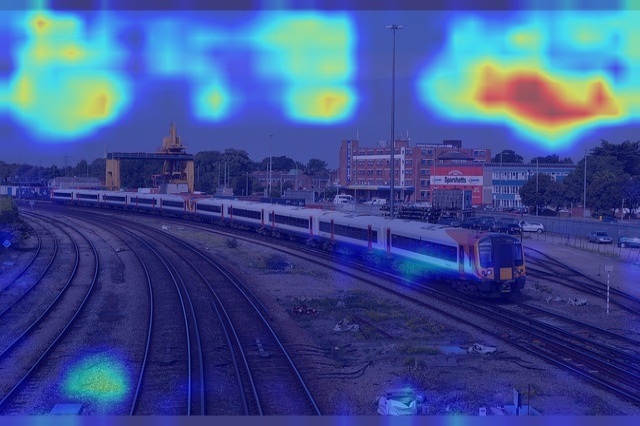}}
    \vfill
    \subfigure[]
    {\label{fig:9c}\includegraphics[width=0.33\linewidth]{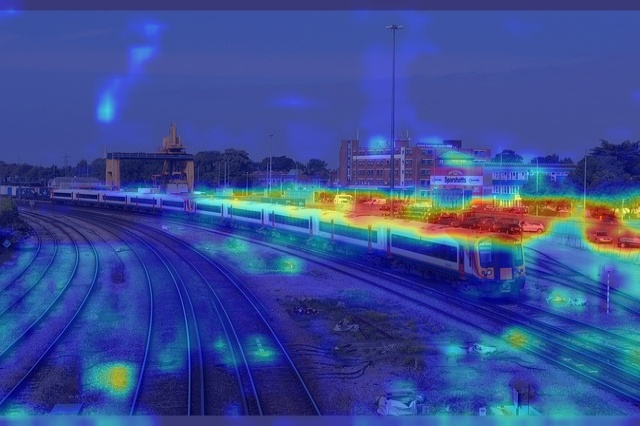}}
    \hspace{0.05em}
    \subfigure[]
    {\label{fig:9d}\includegraphics[width=0.33\linewidth]{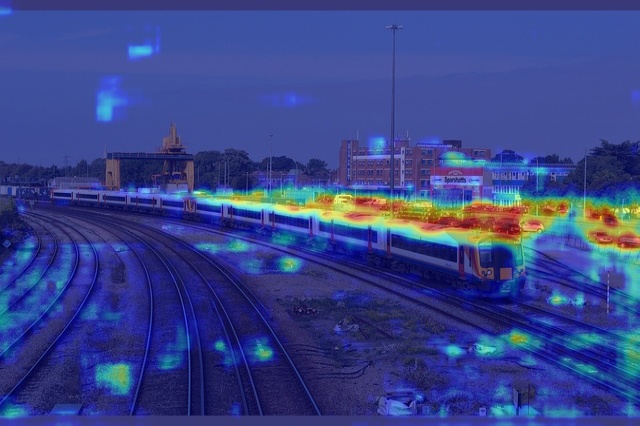}}
\end{center}
\caption{Output heatmaps of YOLOv5s-Tiny neck network.}
\label{fig:9}
\end{figure*}

By incorporating tiny object detection layers in the neck network, YOLOv5s-Tiny outperforms the baseline YOLOv5s in several metrics, as detailed in Table \ref{tab:2}. This enhancement has led to a 1\% gain in Precision$_{all}$, a 2.2\% gain in Recall$_{all}$, a 1.8\% gain in mAP@0.50, and a 1.4\% gain in mAP@0.5:0.95. However, this has resulted in higher computational and parameter requirements, likely due to the additional convolutional modules in the neck network that create larger feature maps.

To highlight the efficacy of the proposed tiny object detection layer, we utilize result images and head network heat maps from both the YOLOv5s and YOLOv5s-Tiny models. Fig.\ref{fig:7} shows (a) the original image, (b) detection results of YOLOv5s, and (c) detection results of YOLOv5s-Tiny, indicating a superior ability of YOLOv5s-Tiny in detecting small objects. Fig.\ref{fig:8} presents three feature heat maps from the YOLOv5s neck network, corresponding to the detection layers for large, medium, and small objects, while Fig.\ref{fig:9} illustrates the corresponding layers from YOLOv5s-Tiny, including the tiny object detection layer. Notably, Figs.\ref{fig:9c} and \ref{fig:9d} reveal that the tiny object detection layer in YOLOv5s-Tiny offers more detailed attention to small objects.

\subsection{Evaluations on lightweight convolution modules}

In this study, four methods to lighten the YOLOv5 model are proposed. The first method, termed YOLOv5s-G1, replaces the C3 module in the backbone with a C3Ghost module. The second method, YOLOv5s-G2, switches out all C3 modules in the network for C3Ghost modules. The third method, YOLOv5s-CC1, swaps the C3 module in the backbone for a C3CrossCovn module. To avoid confusion, the fourth method, which replaces all network C3 modules with C3CrossCovn modules, is named YOLOv5s-CC2. The experimental results for these four models are compiled in Table \ref{tab:3}.
\begin{table*}[t!]
\centering
\caption{Detection results of four lightweight convolution modules.}
\label{tab:3}
\resizebox{\linewidth}{!}{%
\begin{tabular}{lccccccc}
\toprule
Models & Precision$_{all}$ & Recall$_{all}$ & mAP@0.50 & mAP@0.5:0.95 & F1 & GFLOPs & Parameters (M) \\
\midrule
YOLOv5s~\cite{rev1}     & 67.7\% & 50.3\% & 55.7\% & 36.1\% & 0.577 & 16.6 & 7.23 \\
YOLOv5s-G1  & 65.1\% & 51.4\% & 55.0\% & 35.5\% & 0.574 & 13.0 & 6.06 \\
YOLOv5s-G2  & 65.2\% & 50.8\% & 54.4\% & 35.0\% & 0.571 & \textbf{11.1} & \textbf{5.10} \\
YOLOv5s-CC1 & \textbf{68.0\%} & \textbf{52.6\%} & \textbf{57.6\%} & \textbf{37.7\%} & \textbf{0.593} & 16.2 & 7.03 \\
YOLOv5s-CC2 & 65.0\% & 51.1\% & 55.0\% & 34.5\% & 0.572 & 14.1 & 6.00 \\
\bottomrule
\end{tabular}%
}
\end{table*}

\begin{table*}[t!]
\centering
\caption{Detection results of seven attention mechanisms.}
\label{tab:4}
\resizebox{\linewidth}{!}{%
\begin{tabular}{lccccccc}
\toprule
Models & Precision$_{all}$ & Recall$_{all}$ & mAP@0.50 & mAP@0.5:0.95 & F1 & GFLOPs & Parameters (M) \\
\midrule
YOLOv5s~\cite{rev1}       & 67.7\% & 50.3\% & 55.7\% & 36.1\% & 0.577 & 16.6 & \textbf{7.23} \\
YOLOv5s+CA    & 67.9\% & 53.2\% & 57.7\% & 37.6\% & 0.597 & 16.5 & 7.26 \\
YOLOv5s+CBMA  & 66.7\% & 53.1\% & 57.2\% & 37.0\% & 0.591 & 16.5 & 7.27 \\
YOLOv5s+SENet & 65.8\% & 52.9\% & 56.9\% & 36.8\% & 0.586 & 16.5 & 7.27 \\
YOLOv5s+SA    & 68.4\% & 51.9\% & 57.0\% & 36.9\% & 0.590 & \textbf{16.4} & \textbf{7.23} \\
YOLOv5s+ECA   & 66.1\% & 53.5\% & 57.4\% & 37.1\% & 0.591 & \textbf{16.4} & \textbf{7.23} \\
YOLOv5s+GAM   & \textbf{70.0\%} & 53.9\% & 59.2\% & 39.4\% & 0.609 & 22.0 & 9.54 \\
YOLOv5s+SKNet & 68.8\% & \textbf{55.3\%} & \textbf{60.2\%} & \textbf{40.1\%} & \textbf{0.613} & 87.1 & 36.64 \\
\bottomrule
\end{tabular}%
}
\end{table*}

\begin{figure}[t!]
    \begin{center}
        \includegraphics[width=\linewidth]{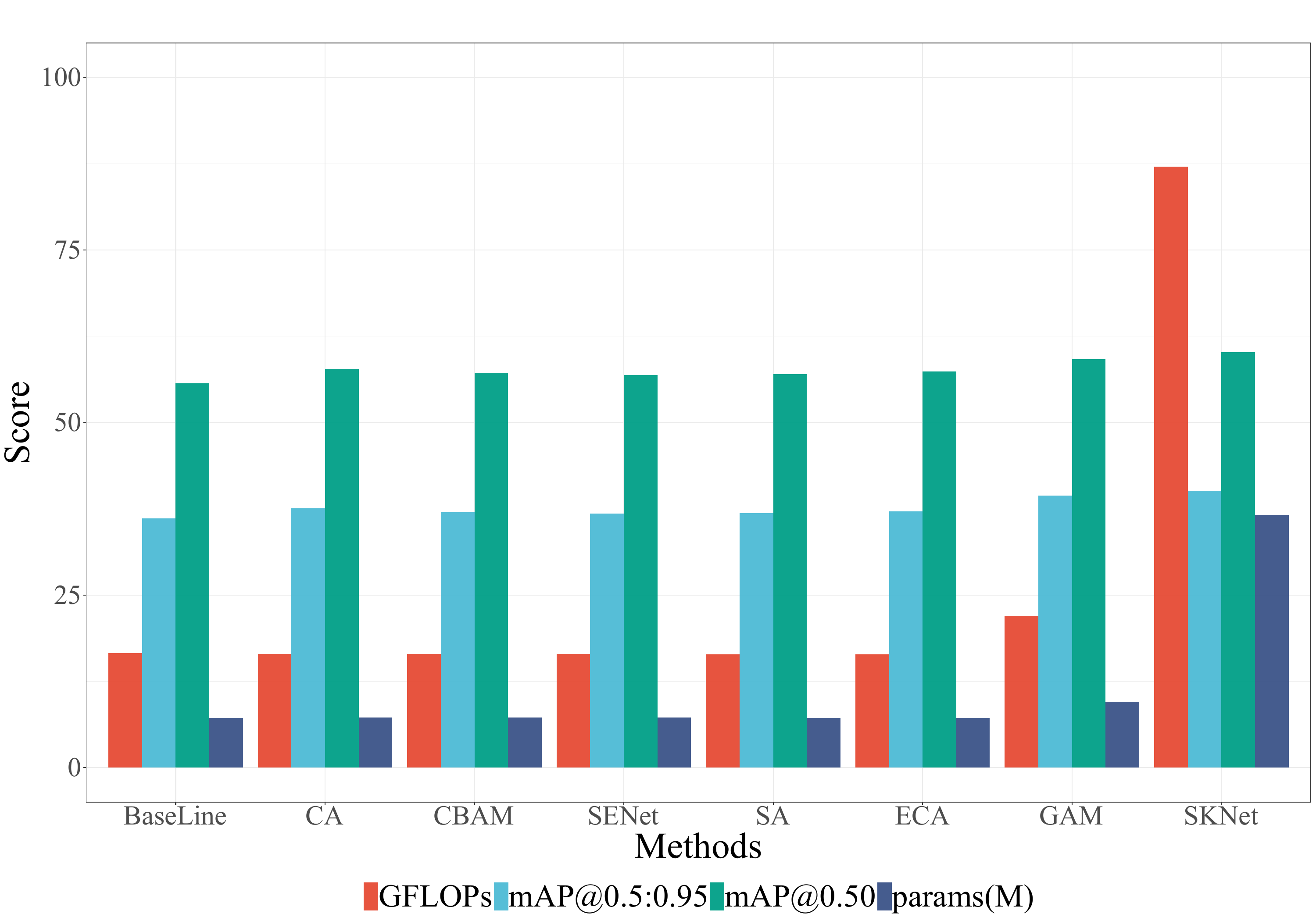}
    \end{center}
    \caption{Detection performance of seven attention mechanisms on YOLOv5s.}
    \label{fig:10}
\end{figure}

The four lightweight improved models, particularly YOLOv5s-G1 and YOLOv5s-G2, have achieved reductions in parameters and computational demands compared to the baseline YOLOv5 model. However, YOLOv5s-G1 and YOLOv5s-G2 also experienced a notable decrease in detection accuracy. Specifically, compared to the baseline model, YOLOv5s-G1 decreased the mAP@0.50 by 0.7\%, mAP@0.5:0.95 by 0.6\%, the number of parameters by 1.17M, and the computational complexity by 3.6GFLOPs. For the YOLOv5s-G2, its mAP@0.50 value decreased by 1.3\% compared to YOLOv5s, mAP@0.5:0.95 decreased by 1.1\%, the number of parameters decreased by 2.13M, and the computational complexity decreased by 5.5GFLOPs. These drops can be attributed to the loss of channel features due to the feature map splicing and the varying receptive fields of GhostCovn, leading to diminished representation capability. The C3CrossConv module includes CrossConv, which focuses more on the main regions of an image, often containing more detectable objects while selectively ignoring other regions with fewer objects. This feature extraction method somewhat reduces model complexity and computational cost without sacrificing accuracy. On a positive note, the YOLOv5s-CC1 model not only reduced its parameter count by 0.2M and computational complexity by 0.4GFLOPs but also improved accuracy, with a 1.9\% increase in mAP@0.50 and a 1.6\% increase in mAP@0.5:0.95. These results indicate the efficacy of the C3CrossCovn module in enhancing backbone network feature extraction, although it presents challenges for multi-scale feature fusion in the neck network and impacts accuracy due to the overall reduction in model parameters and complexity.

In summary, the application of C3CrossConv to the backbone network yielded gratifying results, achieving a dual effect: it not only reduces the number of parameters and computational cost of the model but also enhances the accuracy and efficiency. Therefore, we choose to apply C3CrossConv in the backbone network of YOLO-TLA to achieve optimization.

\subsection{Evaluations on different attention mechanisms}

YOLO-TLA employs GAM within its backbone network and compares it with other prevalent attention mechanisms such as CA, CBAM, and ECA in terms of their effects on detection performance. As detailed in Table \ref{tab:4}, each attention mechanism enhances various detection metrics of YOLOv5s to different extents. The YOLOv5s model with SKNet demonstrates a significant increase of 4.5\% in mAP@0.50 and 4\% in mAP@0.5:0.95, albeit at the cost of a nearly fivefold increase in model parameters and complexity, conflicting with the lightweight objective of this study. Conversely, the YOLOv5s model integrated with GAM achieves improvements of 3.5\% in mAP@0.50 and 3.3\% in mAP@0.5:0.95, with minimal impact on model size and complexity. Therefore, among the tested attention mechanisms, GAM stands out as the most effective in enhancing detection accuracy without substantially increasing model size, maintaining an optimal balance. The comparative impacts of these attention mechanisms on YOLOv5s are visually presented in Fig.\ref{fig:10}, using metrics like mAP@0.50, mAP@0.5:0.95, parameter count, and computational complexity.

\subsection{Discussions}

To illustrate the effects of each enhancement on model performance, the most effective improvement method from each category is chosen in this section. Specifically, YOLOv5s-TL adds both a tiny object layer and a lightweight improvement, based on the YOLOv5s-CC1 model. YOLOv5s-TA combines a tiny object detection layer with GAM but retains the standard C3 module. Finally, YOLO-TLAs, the model proposed in this paper, integrates the tiny object detection layer, the lightweight C3CrossCovn module, and GAM. The outcomes of these various modifications are detailed in Table \ref{tab:5}.

\begin{table*}[t!]
\centering
\caption{Detection results of different improved models.}
\label{tab:5}
\resizebox{\linewidth}{!}{%
\begin{tabular}{lccccccc}
\toprule
Models & Precision$_{all}$ & Recall$_{all}$ & mAP@0.50 & mAP@0.5:0.95 & F1 & GFLOPs & Parameters (M) \\
\midrule
YOLOv5s~\cite{rev1}       & 67.7\% & 50.3\% & 55.7\% & 36.1\% & 0.577 & \textbf{16.6} & 7.23 \\
YOLOv5s-Tiny  & 68.7\% & 52.5\% & 57.5\% & 37.5\% & 0.595 & 19.9 & 7.38 \\
YOLOv5s-TL    & 67.0\% & 53.7\% & 58.1\% & 38.0\% & 0.596 & 19.7 & \textbf{7.19} \\
YOLOv5s-TA    & 70.4\% & 53.9\% & 59.6\% & 39.7\% & 0.611 & 25.5 & 9.70 \\
YOLO-TLA (Ours)& \textbf{71.2\%} & \textbf{57.3\%} & \textbf{60.3\%} & \textbf{40.1\%} & \textbf{0.635} & 25.3 & 9.49 \\
\bottomrule
\end{tabular}%
}
\end{table*}

\begin{table*}[t!]
\centering
\caption{Detection results of different model sizes.}
\label{tab:6}
\resizebox{\linewidth}{!}{%
\begin{tabular}{lccccccc}
\toprule
Models & Precision$_{all}$ & Recall$_{all}$ & mAP@0.50 & mAP@0.5:0.95 & F1 & GFLOPs & Parameters (M) \\
\midrule
YOLOv5s~\cite{rev1}       & 67.7\% & 50.3\% & 55.7\% & 36.1\% & 0.577 & \textbf{16.6} & \textbf{7.23} \\
YOLOv-TLAs (Ours)   & \textbf{71.2\%} & \textbf{57.3\%} & \textbf{60.3\%} & \textbf{40.1\%} & \textbf{0.635} & 25.3 & 9.49 \\
\hline
YOLOv5m~\cite{rev1}       & 70.7\% & 58.2\% & 63.1\% & 43.4\% & 0.638 & \textbf{48.9} & \textbf{21.17} \\
YOLOv-TLAm (Ours)    & \textbf{74.6\%} & \textbf{60.3\%} & \textbf{64.8\%} & \textbf{45.3\%} & \textbf{0.667} & 73.1 & 27.53 \\
\bottomrule
\end{tabular}%
}
\end{table*}

\begin{figure*}[t!]
\begin{center}
    \subfigure
    {\label{fig:11a1}\includegraphics[width=0.248\linewidth]{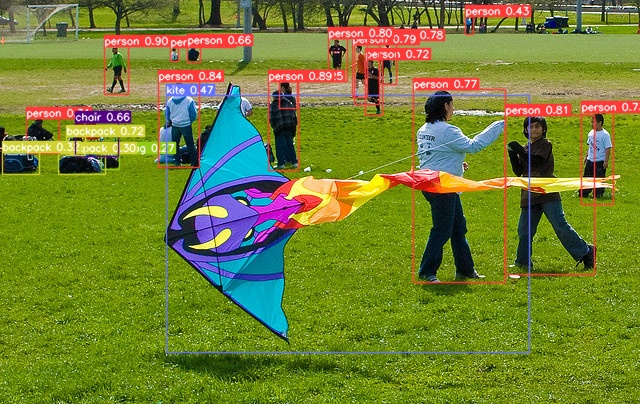}}
    \hfill
    \subfigure
    {\label{fig:11b1}\includegraphics[width=0.248\linewidth]{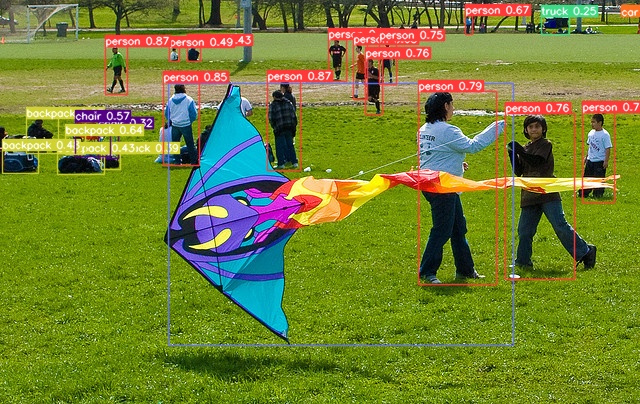}}
    \hfill
    \subfigure
    {\label{fig:11c1}\includegraphics[width=0.248\linewidth]{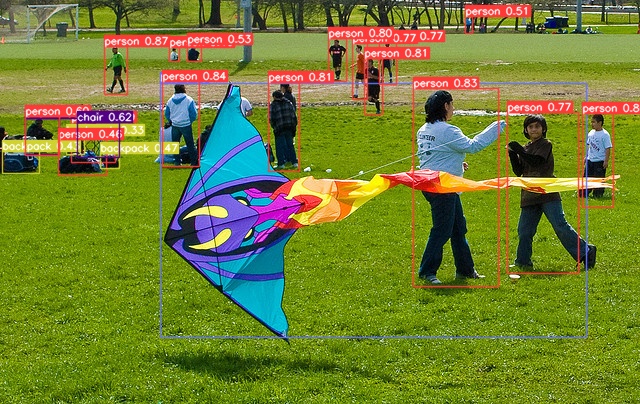}}
    \hfill
    \subfigure
    {\label{fig:11d1}\includegraphics[width=0.248\linewidth]{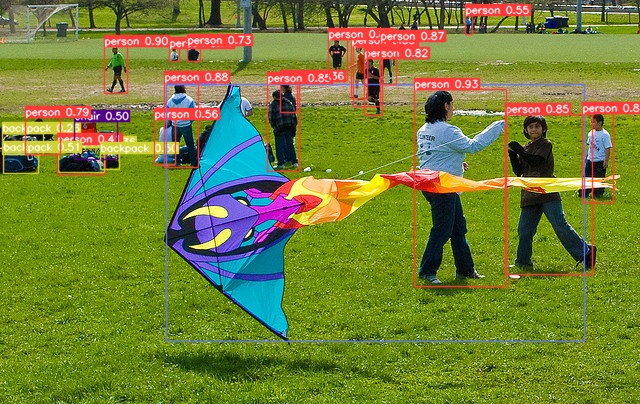}}\vspace{-0.8em}
    \vfill
    \subfigure
    {\label{fig:11a2}\includegraphics[width=0.248\linewidth]{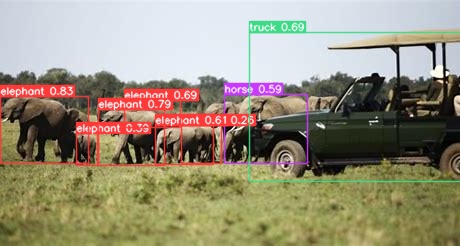}}
    \hfill
    \subfigure
    {\label{fig:11b2}\includegraphics[width=0.248\linewidth]{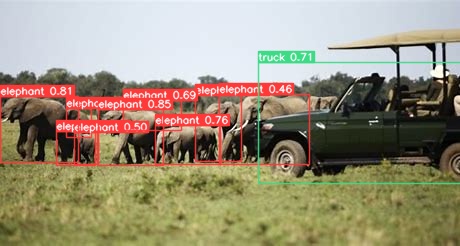}}
    \hfill
    \subfigure
    {\label{fig:11c2}\includegraphics[width=0.248\linewidth]{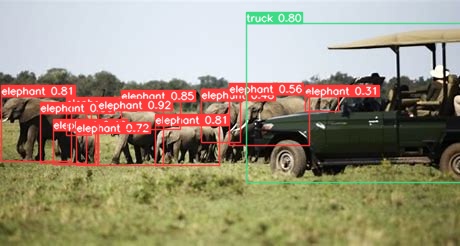}}
    \hfill
    \subfigure
    {\label{fig:11d2}\includegraphics[width=0.248\linewidth]{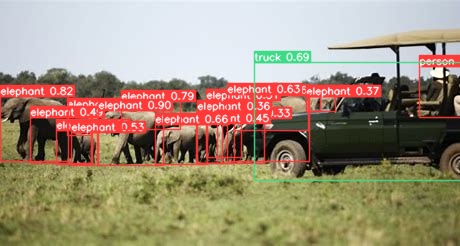}}\vspace{-0.8em}
    \vfill
    \addtocounter{subfigure}{-8}
    \subfigure[YOLOv5s]
    {\label{fig:11a3}\includegraphics[width=0.248\linewidth]{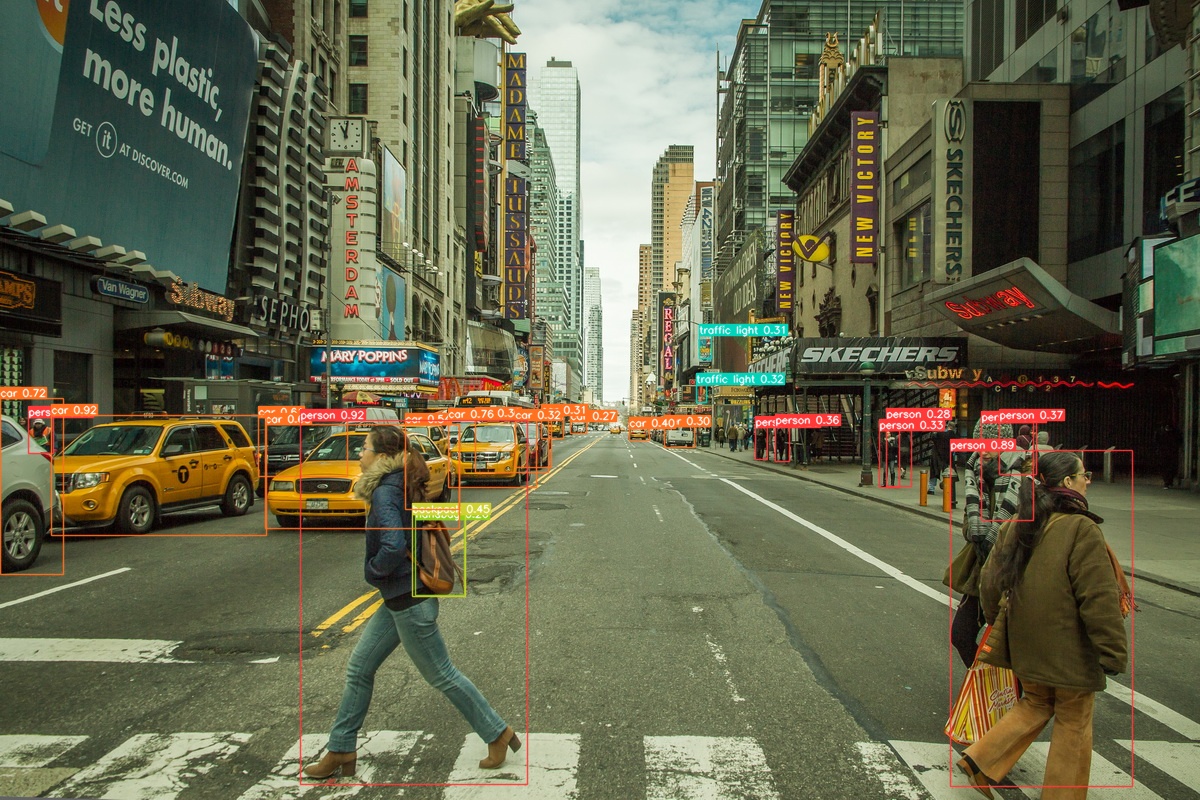}}
    \hfill
    \subfigure[YOLO-TLAs]
    {\label{fig:11b3}\includegraphics[width=0.248\linewidth]{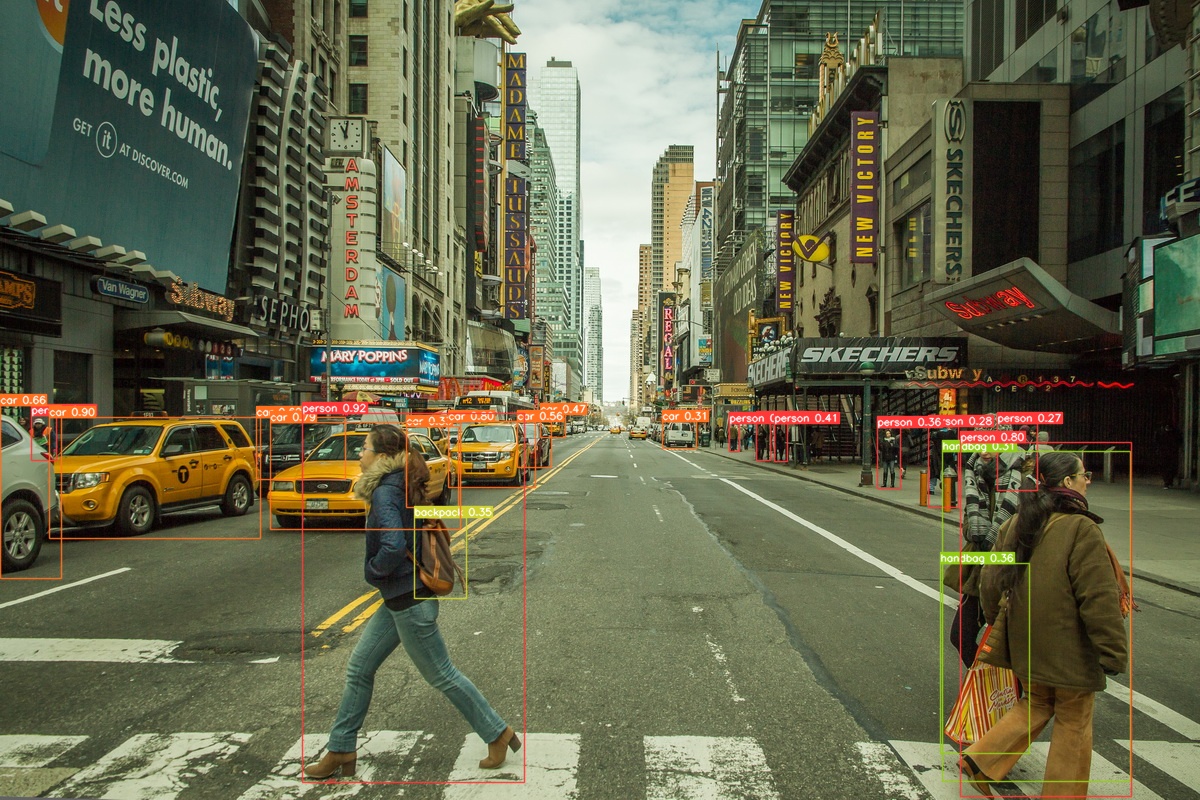}}
    \hfill
    \subfigure[YOLOv5m]
    {\label{fig:11c3}\includegraphics[width=0.248\linewidth]{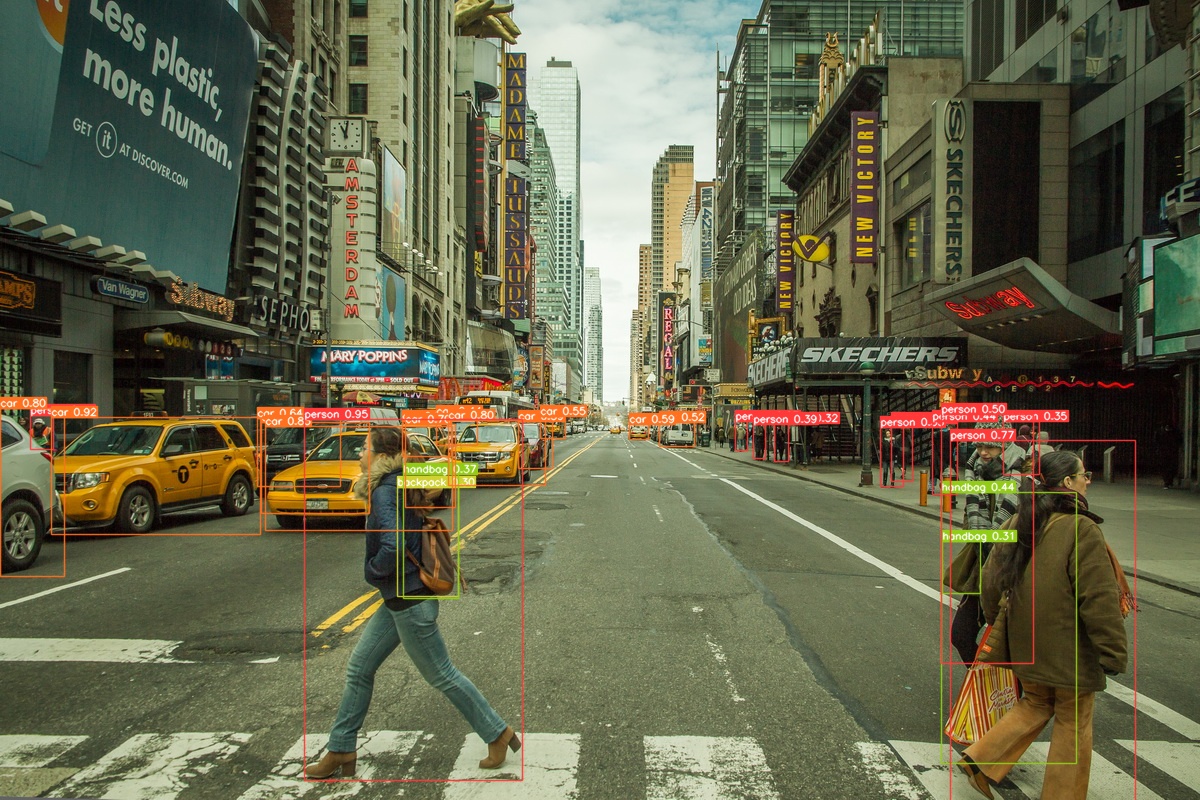}}
    \hfill
    \subfigure[YOLO-TLAm]
    {\label{fig:11d3}\includegraphics[width=0.248\linewidth]{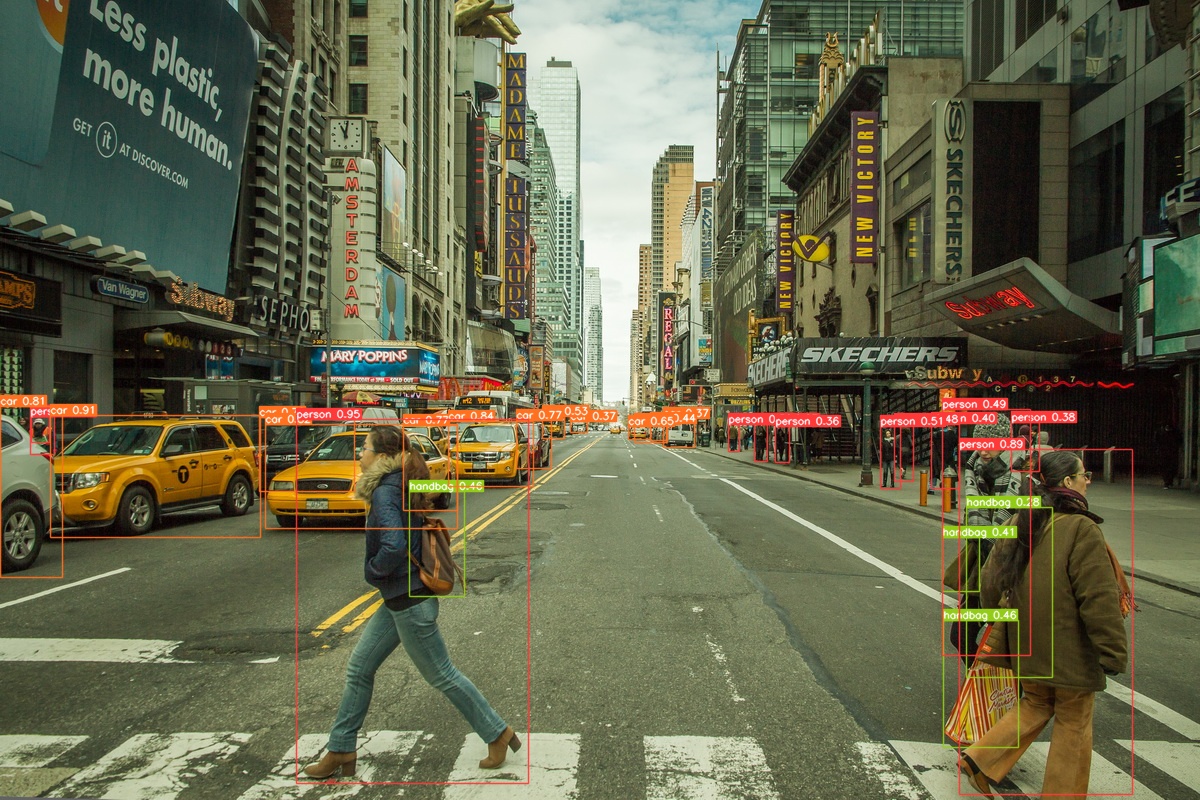}}
\end{center}
\caption{Visualized results of different detection models.}
\label{fig:11}
\end{figure*}

\begin{table*}[t!]
\centering
\caption{Comparative results of YOLO-TLAs against other lightweight object detection models.}
\label{tab:7}
\begin{tabular}{lccc}
\toprule
Models & mAP@0.5:0.95 & GFLOPs & Parameters (M) \\
\midrule
YOLOv5s~\cite{rev1}       & 36.1\% & 16.6 & 7.23 \\
YOLOx-s~\cite{r11}       & 39.8\% & 26.8 & 9.00 \\
YOLOv6n~\cite{r12}       & 37.2\% & 11.4 & 4.70 \\
YOLOv8n~\cite{rev5}       & 36.9\% &  8.7 & \textbf{3.20} \\
EfficientDet-D1~\cite{r22}& 32.6\% & \textbf{6.1} & 6.60 \\
YOLO-TLAs (Ours)    & \textbf{40.1\%} & 25.3 & 9.49 \\
\bottomrule
\end{tabular}%
\end{table*}

\begin{table*}[t!]
\centering
\caption{Comparative results of YOLO-TLAm against other large-scale object detection models.}
\label{tab:8}
\begin{tabular}{lccc}
\toprule
Models & mAP@0.5:0.95 & GFLOPs & Parameters (M) \\
\midrule
YOLOv5m~\cite{rev1}       & 43.4\% & \textbf{48.9} & \textbf{21.17} \\
YOLOx-m~\cite{r11}       & 44.9\% & 73.8 & 25.30 \\
DERT~\cite{r26}          & 40.6\% & 86.0 & 41.00 \\
YOLO-TLAm (Ours)     & \textbf{45.3\%} & 73.1 & 27.53 \\
\bottomrule
\end{tabular}%
\end{table*}

Compared to YOLOv5s-Tiny, YOLOv5s-TL shows a 0.6\% improvement in mAP@0.50 and a 0.5\% improvement in mAP@0.5:0.95, along with a reduction of 0.19M in parameters. YOLOv5s-TA demonstrates a 2.1\% higher mAP@0.50 and a 2.2\% higher mAP@0.5:0.95 compared to YOLOv5s-Tiny. Relative to YOLOv5s-TA, YOLO-TLAs reduces parameters by 0.21M and computational demands by 0.2GFLOPs, while achieving a 0.7\% improvement in mAP@0.50 and a 0.4\% improvement in mAP@0.5:0.95. These results reveal that the lightweight approach not only lowers model complexity but also enhances detection performance. While improvements in accuracy often lead to increased computational and parameter requirements, the lightweight strategy implemented here effectively limits these increases. The experimental results validate that each proposed improvement positively contributes to the enhanced performance of the YOLOv5s model and confirm that these improvements are not mutually exclusive but rather complementary.

YOLO-TLAs demonstrates notable performance enhancements over YOLOv5s, with increases of 4.6\% in mAP@0.50 and 4\% in mAP@0.5:0.95, and a parameter increase of just 2.26M. Applying the same improvement method to the larger YOLOv5m model resulted in the YOLO-TLAm model. This model shows significant gains, with a 1.7\% increase in mAP@0.50 and a 1.9\% increase in mAP@0.5:0.95, compared to YOLOv5m, but at the cost of an additional 6.36M parameters. YOLOv5 adjusts model size by modifying network dimensions, particularly by varying the repetition of modules. Larger models like YOLO-TLAm generally see more substantial parameter and computational increases compared to similar-sized YOLOv5 models, without corresponding large gains in accuracy. Table \ref{tab:6} presents a performance comparison of different YOLO-TLA models with their YOLOv5 equivalents.

To demonstrate the efficacy of the YOLO-TLA method, we compare it with YOLOv5 using three test images randomly taken from different validation sets, characterized by densely populated and uniformly sized objects. Fig.\ref{fig:11} reveals that YOLO-TLAs outperforms YOLOv5s in detecting small objects, while YOLO-TLAm exhibits more comprehensive detection capabilities than YOLOv5m. Overall, YOLO-TLA demonstrates superior detection accuracy and robustness.

\subsection{Comparison with state-of-the-arts}

We also compared the performance of the two YOLO-TLA models with other state-of-the-art object detection models. To ensure a comprehensive and fair comparison, we divided the comparisons into two parts based on the number of parameters: YOLO-TLAs versus other lightweight models and YOLO-TLAm versus other large-scale models. The experimental results are presented in Tables \ref{tab:7} and \ref{tab:8}, respectively.

The mAP@0.5:0.95 of YOLO-TLAs increased by 0.3\% compared to YOLOx-s, with a reduction in computation by 1.5 GFLOPs, though the number of parameters increased by 0.49M. Compared to YOLOv5s, the mAP@0.5:0.95 of YOLO-TLAs improved by 4\%, with an increase of only 2.26M in parameters. The mAP@0.5:0.95 of YOLO-TLAs is 7.5\% higher than that of EfficientDet-D1. Although EfficientDet-D1 has a lightweight architecture, the reduction in model complexity has impacted accuracy.

YOLO-TLAm also performs better than YOLOx-m of the same scale, with a mAP@0.5:0.95 increase of 0.4\% and a decrease in computation by 0.7 GFLOPs. Additionally, compared to the DETR model, although the mAP@0.5:0.95 of YOLO-TLAm only improved by 4.7\%, its computation and parameter count are significantly lower than DETR.

\section{Conclusion}\label{sec:5}

This study tackles prevalent challenges in object detection communities and introduces a novel method, YOLO-TLA. This method demonstrates superior detection performance compared to the baseline YOLOv5, particularly in accurately identifying small objects. YOLO-TLA enhances YOLOv5 by integrating a tiny object detection layer into its neck network, and by incorporating a global attention mechanism to increase accuracy. To balance enhanced detection with model efficiency, a lightweight strategy is employed, incorporating the C3CrossCovn module in the backbone network to decrease model complexity. This strategy not only lowers complexity but also boosts accuracy, as confirmed by experimental results. Furthermore, we also successfully apply these improvements to the larger YOLOv5m model, creating the YOLO-TLAm model, which outperforms YOLOv5m in accuracy and stability. The results validate that the proposed enhancements are effective for larger models as well.

\bibliographystyle{num}
\bibliography{bibliography}

\end{document}